# TOD-ProcBench: Benchmarking Complex Instruction-Following in Task-Oriented Dialogues

**Sarik Ghazarian**[1] *   **Abhinav Gullapalli**[1] *†   **Swair Shah**[1] ‡   **Anurag Beniwal**[1] ‡

**Nanyun Peng**[1]   **Narayanan Sadagopan**[1]   **Zhou Yu**[1] †

[1]Amazon

{sghazari, gabhik, shahswai, beanurag, pengnany, sdgpn, amznzya}@amazon.com

## Abstract

In real-world task-oriented dialogue (TOD) settings such as customer support for trip booking, banking, and healthcare, agents are required to strictly adhere to complex instructions while conducting multi-turn conversations with customers. These instructions are typically presented in natural language format and include general guidelines and step-by-step procedures with complex constraints. Existing TOD benchmarks often oversimplify the complex nature of these instructions by reducing them to simple schemas composed of intents, slots, and API call configurations. To address this gap and systematically benchmark LLMs' instruction-following capabilities, we propose **TOD-ProcBench**, a challenging *benchmark* featuring complex *process* instructions with intricate, fine-grained constraints that evaluates various LLMs' abilities to understand and follow instructions in *multi-turn TODs*. Our benchmark dataset comprises instruction documents derived from the high-quality ABCD dataset with corresponding conversations under human quality control. We formulate fine-grained constraints and action procedures as multi-level condition-action instruction statements. We design three tasks to comprehensively benchmark LLMs' complex instruction-following capabilities in multi-turn TODs. Task 1 evaluates how LLMs retrieve the most relevant statement from a complex instruction and predict the corresponding next action. In Task 2, we synthesize instruction-violating responses by injecting inconsistencies and manipulating the original instructions, and then we analyze how effectively LLMs can identify instruction-violating responses. Task 3 investigates LLMs' abilities in conditional generation of instruction-following responses based on the original complex instructions. Our benchmarking results reveal significant gaps in LLMs' instruction-following capabilities in multi-turn conversations across all three tasks. Additionally, we conduct studies on the impact of multilingual settings and different instruction text formats on compliance performance. We release our benchmark under the Llama 3.3 Community License Agreement to advance research on multi-turn TODs' complex instruction-following capabilities. The dataset is available for download at https://www.amazon.science/publications/tod-procbench-benchmarking-complex-instruction-following-in-task-oriented-dialogues.

---

*Equal contribution
†Corresponding author
‡Work performed while at Amazon

# 1 Introduction

Task-oriented dialogue systems (ToDs) take decisive steps and actions when interacting with users with the primary goal of satisfying customer requests [Wei et al., 2018, Sekulic et al., 2024, Xu et al., 2024]. In multi-turn interactions with users, they initially comprehend a user's query and then collect information needed to address the request while following the constraints defined by complex instructions. Complex instructions comprise strict guidelines that dictate under what conditions the actions and API calls must be executed [Rastogi et al., 2020, Raimondo et al., 2023, Chen et al., 2021, Roy et al., 2024]. The complexity of these instructions varies based on the user's intent and the task domain.

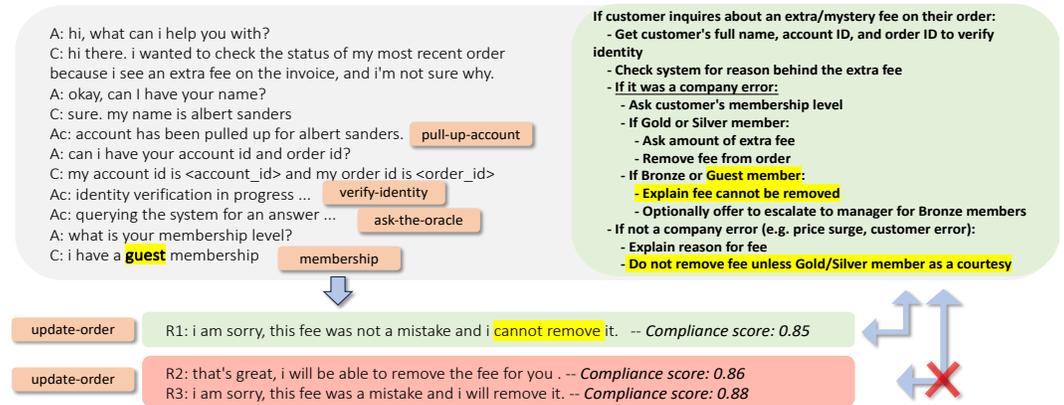

Figure 1: Workflow steps (orange boxes) from ABCD dataset show the the order of calling actions for "status_mystery_fee" intent, an inadequate representation of all nuanced constraints in the corresponding TOD-ProcBench instruction document excerpt on the right. The compliance scores are computed with ComplianceScorer.

Existing datasets either fail to capture challenges due to multi-turn conversation or fail to capture real-world instruction complexity, lacking fine-grained conditional instructions shown in Figure 1. Real-world instructions typically use natural language **IF (Condition) - THEN (Action)** structures which introduce ambiguity compared to symbolic formats but express more complex guidance. This creates a need for public datasets featuring multi-turn conversations that comply with fine-grained natural language conditional instructions. Figure 1 demonstrates how ComplianceScorer [Min et al., 2023] incorrectly labels all responses as compliant when responses R2 and R3 actually violate fee removal instructions, highlighting the need for fine-grained conditional instructions beyond basic workflows. Natural language instructions offer the additional benefit of seamlessly sharing consistent instructions between human agents and chatbots without conversion. For this purpose, we construct a multilingual dataset called TOD-ProcBench which contains conversations and complex instructions in the condition-action format. Our contributions are:

- We propose TOD-ProcBench, based on ABCD [Chen et al., 2021], as a collection of benchmarks with complex condition-action natural language instructions and multilingual multi-turn conversations.

- We publish benchmarks for three TOD tasks: (1) instruction and next action selection - this task measures how well LLMs can identify applicable instructions during a multi-turn conversation (2) instruction-following evaluation - this task measures how well LLMs can detect violation of complex instruction (3) instruction-following response generation - this task measures how well LLMs can respond under complex instructions written for an entire conversational interaction.

- We propose an approach for generating a test benchmark of instruction-violating responses in TOD-ProcBench by injecting different perturbations to instructions and prompting the LLM to generate responses which are similar to the ground-truth yet adhere to manipulated instructions.

- We show that all LLMs considered in this study struggle to perform accurately on (1) complex formatted instruction selection, (2) instruction-violation response evaluation, and (3) instruction-following response generation tasks. In addition, we find that LLMs have similar multilingual capability for those tasks with a subtle preference for English.

## 2  Related Work

In general, ToDs such as customer service chatbots, physical robots, virtual assistants [Cui et al., 2017, Sengupta et al., 2019, Lewis and White, 2023] are strictly required to comply with the predefined procedures when executing API calls and actions to resolve user issues. Datasets have been introduced to incorporate constraints on ToDs through schema-guided paradigms. The schema of the synthetic SGD dataset [Rastogi et al., 2020] imposes constraints through the set of obligatory slots and the order in which they must be filled for actions to be executed. As shown in Appendix A, SGD schema for *finding next flight* intent includes slots (flight's origin and destination, both airports, seating class and etc.), their definitions and values, and required slots for each intent. Similar to SGD, STAR [Mosig et al., 2020] and FLAP [Roy et al., 2024] present the schema concept in ToDs in flowchart format. For these datasets, the schema with actions and slot values do not usually reflect the nature of real-world instructions as the latter often has more complex step-by-step structure and condition-action combination. The Action-Based Conversations Dataset (ABCD) [Chen et al., 2021] which our study is based on incorporates guidelines that a human agent needs to execute in a predefined order. We go a step further and present more realistic, natural language instructions in the form of multi-level condition-action statements that provide precise information about the conditions that are required to be true for actions to be executed. More recent datasets and studies benchmark an LLM agent's ability to follow SOPs [Nandi et al., 2025, Li et al., 2025]. Their focus is more on the accuracy of tool selection alone, complementing our study of conversational capabilities in a multi-turn dialogue setting.

## 3  TOD-ProcBench

### 3.1  Problem Overview

In our setting, to generate a response $r$ for a given dialogue history $u$ (for example, "user: i need to return this jacket. it is the wrong size") within a conversation in language $l$, a TOD system $D$ must follow a set of conditions $\mathbf{c} \subseteq C$ and execute a sequence of corresponding actions $\mathbf{a} = \langle a_1, a_2, \ldots, a_n \rangle$ with $a_i \in A$ outlined in a given instruction $n \in N$. We present multilingual TOD-ProcBench with two main dimensions having three and seven variations, respectively: (1) the format dimension $f$ of $n \in N$ with English natural language ($f_1$="Nested If-Then", $f_2$="Flattened If-Then", and $f_3$="Flattened JSON Mappings of Conditions and Actions") and (2) the language dimension $l$ of the conversation (Arabic, Chinese, English, French, German, Hindi, Spanish).

### 3.2  Complex Instruction Components

#### 3.2.1  Conditions

Specific states or attributes, such as customer attributes (e.g., membership level), item properties (e.g., purchase date), or conversation states (e.g., user intent) must be verified before executing a sequence of corresponding actions.

#### 3.2.2  Actions

The TOD must execute specific operations or responses when associated conditions are satisfied. Actions are organized hierarchically with conditions. For example, when a user inquires about the return policy (1st-level condition), the agent must first identify the user's membership level (1st-level action). If the user has a *silver* membership (2nd-level condition), then the agent provides specific information like "only purchases in the last 6 months are returnable" (2nd-level action).

### 3.2.3 Composition

Similar to prior works such as Wen et al. [2024] and Li et al. [2025], we use six composition techniques to derive complete, complex instructions with the foundational *condition* and *action* components: *Single*, *And*, *Or*, *Chain*, *Selection*, *Nesting*.

### 3.3 Instruction Document Construction

We derive TOD-ProcBench from the ABCD dataset [Chen et al., 2021], a collection of human-written conversations created by annotators following predefined workflow actions. We aggregate all human-human conversations with the same user intent from ABCD, which inherently contain similar conversation trajectories. In Appendix B, we show examples and symbolic notations for each of the three complex instruction formats, $f_1$, $f_2$, $f_3$.

**Format $f_1$** We generated "Nested If-Then" complex instructions, with one distinct instruction document for each of the 55 user intents in ABCD, using LLMs with 1-shot examples for natural language condition-action instructions (averaging 146 conversations per intent). We quantify the complexity of TOD-ProcBench with characterizations such as an average conditional branching factor of 2.2 and up to 4 nested conditional levels. To answer the question "How does the format of complex natural language instructions affect an LLM's compliance capability?", we use $f_1$ to derive two additional formats of the instructions which still present the fundamental *conditions* and *actions* in English natural language text.

**Format $f_2$** "Flattened If-Then" effectively flattens the nested "If-Then" structure of $f_1$ into a sequence of single-level *condition-action* "If-Then" statements. We use this format to evaluate whether LLMs have a strong preference for the *Nesting* and *Chain* composition techniques present in $f_1$ but not $f_2$. The flattened structure of $f_2$ reduces a *Chain* into a series of composite conditions with *And*.

**Format $f_3$** "Flattened JSON Mappings of Conditions and Actions" transforms the natural language "If-Then" statements in $f_2$ into a structured JSON object containing a list of maps, each of which denotes relevant *conditions* and *actions* to be considered.

### 3.4 Expanding to Multilingual Conversations

As a practical application of our work in constructing TOD-ProcBench and a first step to evaluate the multilingual capabilities of LLMs in complex instruction-following, we seek to answer the question "Are English instructions still effective to guide multi-turn conversations in another language?". This study will help justify whether there is a need to produce a high-quality construction of instructions in other languages for production settings where English instructions may be used in global marketplaces. We translate only the multi-turn conversations component of the original ABCD dataset to six other diverse languages: Arabic, Chinese, French, German, Hindi, Spanish.

### 3.5 Quality Verification

We ask human annotators to assess the quality of TOD-ProcBench's English conversation pairings with the 55 generated documents in format $f_1$ for the test conversations of ABCD dataset. The annotators answer the following Yes/No questions shown in the user interface in Figure 3 in Appendix A:

**Accuracy**: Does the generated instruction accurately reflect information in the conversation? If the answer is no, the annotator should provide which part of the instruction is incorrect.

**Missing**: Is there any important information from the conversation that is missing from the model-generated instruction? If the answer is yes, the annotator should specify which part from the conversation is not in the instruction.

Three annotators evaluated each conversation-instruction pair (3,012 total annotations). Annotator agreement was complete for 63% of conversations, with majority voting determining "accuracy" (82%) and "missing" (15%) assessments. The top-3 accurate annotations are for the *recover_password*, *recover_username* and *search_results* intents which share similar conversation trajectories, while the bottom-3 belong to the *shirt*, *timing* and *policy* intents which can include a broad range of queries. We only focus on the 769 high-quality pairs. Using English conversation pairings with English instructions in format $f_1$ as a base version of TOD-ProcBench, we expand the dataset along the two dimensions of instruction format and conversation language. For the format and multilingual expansion datasets, we leverage the LLM judge to evaluate the quality of both, with 100% correctness for the format changes and 98.3% to 99.8% for each language variation.

# 4 Tasks

For all experiments across three tasks, three instruction formats, and seven conversation languages, we prompt the following six LLMs to conduct chain-of-thought (CoT) [Wu et al., 2023] by first generating a step-by-step reasoning and then generate the respective output: Qwen3-14B [Yang et al., 2025], Llama3.3-70B [4], Gemma3-27B-IT [5], Claude3.5-Sonnet-V1 [6], Claude3.5-Sonnet-V2 [7], Claude3.7-Sonnet [8]. Experimental results which are not presented in tables in the following sections are presented in Appendix D for completeness.

## 4.1 Task 1: Instruction Retrieval and Next Action Prediction

Task 1 evaluates ToDs' ability to retrieve relevant instructions and predict appropriate next actions conditioned on conversation history. For all 30 actions (details in Appendix A, C) from the ABCD dataset, we define trigger conditions that determine when each action should be executed. We also introduce an *empty* action to handle scenarios in which the customer's goal has already been met and no further action is required from the agent.

### 4.1.1 Dataset

To evaluate model performance in Task 1 at different stages of the user-agent interaction, we create input data by extracting all partial conversation transcripts that end with the customer's turn in a given conversation. For example, a conversation with four customer turns leads to a list of four partial conversations, each ending with a customer turn. We then pair each partial conversation with its corresponding instruction. Overall, 6953 partial conversations are extracted from 1004 test conversations.

In order to examine the accuracy of models' predictions for the agent's next action, we process the list of workflow steps from ABCD for each partial conversation history and identify actions that are most relevant to the existing agent turns in the conversation history. The first action from the remaining workflow steps is used as the ground-truth next agent action. Examples of input data for Task1 are shown in Appendix A.

### 4.1.2 Results

We prompt six LLMs to generate the the next agent turn and top $k = 1$ or $k = 2$ most relevant instruction given the conversation history, list of all instructions and possible agent actions with the conditions for invoking each of these actions. In computing the top $k = 2$ accuracy, we check whether any of the two relevant instructions retrieved by LLMs is the ground-truth instruction or not. All models are executed in a few-shot in-context learning [Brown et al., 2020] paradigm where we provide three demonstrations for LLMs.

Table 1 depicts the accuracy of the models in predicting the correct instruction and agent's next action. While the larger LLMs maintain a relatively more consistent performance for $k = 1$ and $k = 2$, all LLMs recorded higher accuracy for the top-$k = 2$ scenario for both overall accuracy and only instruction retrieval. Most LLMs demonstrate a preference for format $f_1$ and all LLMs demonstrate more difficulty in correctly predicting the agent's next action, likely attributable to the subtle distinctions between actions and the overlapping nature of their triggering conditions. We also compute accuracy of the models in predicting the correct instruction and action based on relative position of the given agent response in the conversation, assigning turns to beginning, middle or end based on whether they are located in the beginning, middle, or final third of the agent turns in the conversation. According to Table 1, larger models perform consistently throughout the conversation. Although smaller models perform better at instruction retrieval and action predictions as they get more information from the conversation history, all LLMs still have more room for improvement. As shown in Appendix D, larger models tend to outperform smaller models across the conversation languages but performance of all LLMs is not severely impacted by language.

---

[4] https://huggingface.co/meta-llama/Llama-3.3-70B-Instruct
[5] https://huggingface.co/google/gemma-3-27b-it
[6] https://www.anthropic.com/news/claude-3-5-sonnet
[7] https://www.anthropic.com/news/3-5-models-and-computer-use
[8] https://www.anthropic.com/news/claude-3-7-sonnet

| Model | Overall | Instruction Retrieval | Begin | Middle | End | Best Format |
|---|---|---|---|---|---|---|
| Claude3.7-Sonnet | **0.3948 / 0.4310** | **0.8267 / 0.8943** | **0.3878 / 0.4481** | **0.3834** / 0.4074 | 0.4161 / 0.4359 | $f_2$ / $f_1$ |
| Claude3.5-Sonnet-V2 | 0.3312 / 0.4182 | 0.7311 / 0.8765 | 0.3270 / 0.4214 | 0.3333 / **0.4126** | 0.3341 / 0.4204 | $f_1$ / $f_3$ |
| Claude3.5-Sonnet-V1 | 0.3187 / 0.3716 | 0.7057 / 0.8304 | 0.3185 / 0.3750 | 0.3342 / 0.3730 | 0.3018 / 0.3660 | $f_3$ / $f_3$ |
| Qwen3-14B | 0.3122 / 0.3279 | 0.7792 / 0.8419 | 0.1645 / 0.1668 | 0.2828 / 0.2880 | **0.5289 / 0.5728** | $f_1$ / $f_1$ |
| Gemma3-27B-IT | 0.2221 / 0.2559 | 0.6138 / 0.7559 | 0.0770 / 0.0755 | 0.2144 / 0.2392 | 0.4113 / 0.4990 | $f_1$ / $f_2$ |
| Llama3.3-70B | 0.1447 / 0.2497 | 0.2503 / 0.6612 | 0.1463 / 0.2620 | 0.1477 / 0.2366 | 0.1393 / 0.2488 | $f_2$ / $f_1$ |

Table 1: Task 1 accuracy and best performing instruction format for top $k = 1$ and $k = 2$ instruction retrieval and next agent action prediction given English conversations. Each entry is given in the form $k = 1/k = 2$. "Overall" denotes both instruction retrieval and next action prediction as opposed to only "Instruction Retrieval". "Begin", "Middle", and "End" denote the respective 1/3 segments of the conversation. The best result per column is **bolded**.

| Language | Claude3.7-Sonnet | Claude3.5-Sonnet-V2 | Claude3.5-Sonnet-V1 | Qwen3-14B | Gemma3-27B-IT | Llama3.3-70B |
|---|---|---|---|---|---|---|
| EN | **0.3959/0.3932** | 0.3270/0.3312 | 0.3019/0.2963 | 0.1645/0.3122 | 0.0770/0.2221 | 0.1374/0.1358 |
| FR | **0.4141/0.4041** | 0.3615/0.3613 | 0.3522/0.3299 | 0.1540/0.2976 | 0.0441/0.1877 | 0.1188/0.1146 |
| DE | **0.3885/0.3896** | 0.3483/0.3524 | 0.3386/0.3283 | 0.1242/0.2786 | 0.0708/0.2088 | 0.1262/0.1126 |
| ES | **0.3955/0.4005** | 0.3518/0.3472 | 0.3282/0.3268 | 0.1521/0.2874 | 0.0658/0.1978 | 0.1327/0.1280 |
| ZH | **0.3808/0.3768** | 0.3212/0.3209 | 0.3185/0.3216 | 0.1521/0.2983 | 0.0550/0.1921 | 0.0851/0.0840 |
| AR | **0.3820/0.3778** | 0.3007/0.3036 | 0.3293/0.3206 | 0.1726/0.2934 | 0.0433/0.1585 | 0.0298/0.0299 |
| HI | **0.3785/0.3882** | 0.3317/0.3141 | 0.3386/0.3365 | 0.1718/0.3017 | 0.0701/0.1878 | 0.0252/0.0240 |

Table 2: Task 1 accuracy for predicting the correct next action and retrieving the correct top $k = 1$ instruction in the beginning 1/3 of the conversation (Begin) and for every agent response (Overall), denoted by Begin/Overall entries in the table. Results are shown across models and conversation languages (English (EN), French (FR), German (DE), Spanish (ES), Chinese (ZH), Arabic (AR), Hindi (HI)). The best result per language is **bolded** and the best result per model is <u>underlined</u>.

## 4.2 Task 2: Compliance Evaluation

The main goal of Task 2 is to benchmark LLMs' capability on scoring the compliance of a given ToD's response with respect to the dialogue history and the instruction that the agent must follow. Such capability is essential to be used as both a quality verifier during inference and a proxy for improving the development process for instruction-following ToDs. We formulate a binary classification task where the goal is to distinguish between *instruction-following* and *instruction-violating* agent responses.

**Dataset** For the compliance evaluation task, we only focus on agent utterances that are directly relevant to instructions. To isolate such responses from generic responses such as greeting / acknowledging customer / pure empathy, we leverage a LLM to classify agent turns as *relevant* or *irrelevant* in the 6953 partial conversations from Task 1 dataset.

**Compliance Manipulation Pipeline** To build our binary instruction-following / instruction-violating dataset, we use human-written ABCD responses as instruction-following examples and generate instruction-violating counterparts through a four-step pipeline (Figure 2): (1) extract relevant instruction sections, (2) manipulate the extracted instructions, (3) generate instruction-violating responses based on the manipulated instructions, (4) verify whether the generated instruction-violating response entails the original ground-truth response.

1) **Instruction Section Extraction**: We use dialogue history, instruction document, and the ground-truth instruction-following agent response as input when prompting the LLM to select a section (including its closest 'if' condition) of the instruction to which the ground-truth response is directly related. If such a relevant part of the instruction is not present, we exclude this response sample.

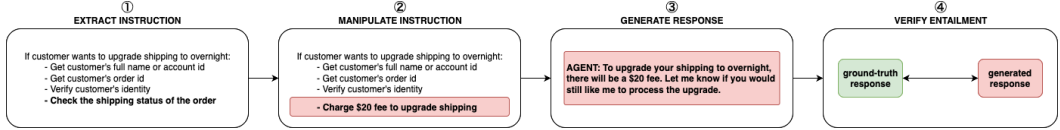

Figure 2: The proposed pipeline for generating instruction-violating responses for Task 2.

| Model | EN | FR | DE | ES | ZH | AR | HI |
|---|---|---|---|---|---|---|---|
| Claude3.7-Sonnet | **0.7608/0.7298** | **0.7558/0.7233** | **0.7533/0.7260** | **0.7490/0.7283** | **0.7407/0.7170** | **0.7510/0.7205** | **0.7472/0.7152** |
| Claude3.5-Sonnet-V2 | 0.7215/0.6874 | <u>0.7253</u>/0.6776 | 0.7187/0.6763 | 0.7207/<u>0.6887</u> | 0.7175/0.6640 | 0.7114/0.6582 | 0.7033/0.6524 |
| Claude3.5-Sonnet-V1 | 0.6847/<u>0.6854</u> | 0.6867/0.6811 | <u>0.6869</u>/0.6784 | 0.6834/0.6796 | 0.6615/0.6604 | 0.6483/0.6551 | 0.6435/0.6496 |
| Llama3-70B | <u>0.7283</u>/<u>0.5636</u> | 0.7273/0.5346 | <u>0.7303</u>/0.5388 | 0.7250/0.5356 | 0.6978/0.5545 | 0.6266/0.4970 | 0.6405/0.5129 |
| Qwen3-14B | <u>0.7046</u>/<u>0.6937</u> | 0.6968/0.6905 | 0.6958/0.6872 | 0.6965/0.6847 | 0.6965/0.6907 | 0.6920/0.6837 | 0.6895/0.6758 |
| Gemma3-27B-IT | <u>0.6968</u>/<u>0.5616</u> | 0.6890/0.5436 | 0.6773/0.5477 | 0.6935/0.5482 | 0.6890/0.5404 | 0.6791/0.5376 | 0.6804/0.5409 |

Table 3: Task 2 accuracy comparison between Approach 1/Approach 2 with instruction format $f_1$ across conversation languages (English (EN), French (FR), German (DE), Spanish (ES), Chinese (ZH), Arabic (AR), Hindi (HI)). The best result per language is **bolded** and the best result per model is <u>underlined</u>.

2) **Instruction Manipulation**: We apply systematic manipulations to both the extracted instruction section and the complete source instruction using an LLM to simulate the following categories of violations:

- **Fine-grained parameter mismatch**: We substitute numerical values (e.g., number of days for which a customer is eligible to return an item) in the extracted section of the instruction with randomly generated numbers, simulating subtle compliance violations.

- **Single-condition action replacement**: For instructions containing one condition with multiple actions, we randomly substitute an original action with either its negation or an alternative action from the same instruction, capturing action-level instruction violations.

- **Multi-condition action replacement**: For instructions with a composition of multiple conditions, we alter the logical condition-action relationships by randomly replacing a target action with an action corresponding to other conditional branches within the same instruction. For example, we could apply standard member return policies for a gold member, violating the true dependence of actions on their corresponding conditions.

3) **Response Generation**: We provide the dialogue history, the manipulated and original instructions, and the ground-truth response to the LLM and prompt it to minimally revise the ground-truth response such that it covers the main difference between the original and manipulated instruction. We explain in the prompt that the generated response must comply with the manipulated instruction. This implicitly guarantees that the generated response is an instruction-violating response.

We used LLM to verify that generated instruction-violating responses do not *entail* ground-truth responses. Of 2288 generated samples, 307 required additional manipulation cycles. Manual verification confirmed 87% accuracy in instruction violations (1982 samples), resulting in a final dataset of 3964 balanced examples (compliant and non-compliant) for Task 2.

**Results** We compare the six LLMs as binary classifiers for instruction violations in TOD-ProcBench using two approaches:

- **Approach 1**: Given the conversation, next agent response, and relevant instruction section, the LLM should classify the agent's next response as compliant or instruction-violating.

- **Approach 2**: LLM first attributes the given agent response to the most relevant instruction and then verifies whether the identified instruction and originally provided ground-truth instruction are *entailing* each other. *Entailment* implies a compliant response.

We leverage accuracy as the evaluation metric for Task 2. As Table 3 indicates, most models perform slightly better for English conversations in both approaches. Table 3 also highlights Approach 2's impact on the compliance evaluation performance as all models demonstrate inferior performance when relying on instruction-level comparisons. Approach 2 increases the complexity of the compliance evaluation task by having the model first generate the relevant instruction based on the given conversation and then verify compliance with the generated instruction. Table 32 in the Appendix also demonstrates that the LLMs are unable to conduct the compliance evaluation task accurately and have a subtle preference for format $f_1$, calling for further studies.

### 4.3 Task 3: Compliant Response Generation

Inspired by real-world scenarios of conditional generations [Chen et al., 2019, Siddique et al., 2022, Narayan et al., 2023], in Task 3, we investigate ToDs' conditional generation abilities based on our predefined instructions which require responses to follow the specified guidelines.

**Dataset**   In Task 3, we only examine TODs' abilities to generate follow-up responses given that the previous conversation is already compliant with given instructions. We use the 3310 partial conversations from the first step of the Task2 pipeline containing instruction-following responses, dialogue histories, and corresponding instructions.

**Results**   We instruct the six LLMs to continue the conversation while following the given instruction. In the prompt, we decompose the problem to identify the steps the agent has followed, select the next most relevant action based on the given instruction, and generate a response that matches the instruction's retrieved action. We employ Claude3.7-Sonnet (the top-performing Task 2 LLM) as a judge to assess response quality and compliance without ground-truth knowledge. Using its compliance classifications, we calculate compliance rates for Task 3, where higher scores indicate better consistent adherence to instructions.

| Model | EN | FR | DE | ES | ZH | AR | HI |
|---|---|---|---|---|---|---|---|
| Claude3.7-Sonnet | 0.9529 | **0.9553** | **0.9535** | **0.9628** | **0.9616** | **0.9680** | **0.9622** |
| Claude3.5-Sonnet-V2 | **0.9532** | 0.9505 | 0.9432 | 0.9544 | 0.9529 | 0.9559 | 0.9505 |
| Claude3.5-Sonnet-V1 | 0.9254 | 0.9353 | 0.9344 | 0.9378 | 0.9375 | 0.9429 | 0.9390 |
| Gemma3-27B-IT | 0.8952 | 0.3130 | 0.3121 | 0.3009 | 0.3205 | 0.3278 | 0.3323 |
| Llama3.3-70B | 0.1435 | 0.4773 | 0.5773 | 0.3565 | 0.5227 | 0.3931 | 0.3834 |
| Qwen3-14B | 0.2269 | 0.2281 | 0.2254 | 0.2142 | 0.2326 | 0.2369 | 0.2375 |

Table 4: Task 3 compliance rate across models and conversation languages (English (EN), French (FR), German (DE), Spanish (ES), Chinese (ZH), Arabic (AR), Hindi (HI)) with instruction format $f_1$. The best result per language is **bolded** and the best result per model is underlined.

Table 4 compares the quality of the models' generations using prior criteria from the given instructions. Larger LLMs tend to exhibit higher compliance in comparison to smaller models, across the conversation language and instruction format dimensions. Some smaller models tend to not respond in the specified response format, leading to many non-compliant responses and very low compliance rates. The performance of all LLMs is not significantly impacted by the language. We observe that the models usually generate longer utterances as it is hard for them to accurately identify one next action from the instruction and to exclusively cover that action. Even though the instruction-following generations may not be semantically close to ground-truth responses, Claude3.7-Sonnet still considers them as compliant.

## 5   Conclusion

In this work, we apply complex condition-action formatted natural language instructions for ToDs to closely model the complexity of real world complex instructions. With these instruction documents and their application to multilingual conversations, we propose three related tasks: instruction retrieval and next action prediction, instruction-following evaluation, and compliant response generation. We publish benchmarks for each task in TOD-ProcBench and show that advanced LLMs still strive to perform admirably in all of these tasks and with enhanced multilingual capabilities for more challenging, fine-grained tasks. As future works, we plan to extend TOD-ProcBench to additional dataset formats including codified structures across domain-specific instructions.

### Acknowledgments

We would like to thank Leo Feldman for his invaluable guidance and support in helping us adapt a system that he built for efficient LLM inference.

# Appendix

## A  Datasets

In this section, we present sample data from our proposed TOD-ProcBench and related works.

**SGD (Task: Flights_1)**

description: Find your next flight

slots: [passengers, seating_class, origin_city, destination_city, origin_airport, ...]
    slot: seating_class
        description: Seating class for the booking
        is_categorical: true
        possible_values: [Economy, Premium Economy, Business, First Class]
intents: [SearchOnewayFlight, SearchRoundtripFlights, ReserveOnewayFlight, ReserveRoundtripFlights]
    intent: ReserveOnewayFlight
        description: Reserve a one-way flight
        is_transactional: True
        required_slots: [origin_city, destination_city, airlines, departure_date]
        optional_slots: [refundable, passengers, seating_class]

Table 5: Example of instructions in SGD.

**STAR (Task: book_doctor_appointment)**

graph: Hello → ask_name → ask_doctor_name → ask_day → ask_start_time → ask_symptoms → query_check
                available → inform_booking_available
                unavailable → inform_booking_unavailable
                    yes → query_book
                    no → ask_doctor_name
                        query_book → doctor_inform_booking_successful → inform_booking_successful

Table 6: Example of instructions in STAR.

**ABCD (Task: Policy and Return Due to Size)**

search-faq → search-policy → select-faq
pull-up-account → validate-purchase → membership → enter-details → update-order

Table 7: Example of instructions in ABCD.

**TOD-ProcBench (Task: Policy and Return Due to Size)**

If customer wants to know the return policy:
   - Ask for customer's membership level (gold, silver, bronze, guest)
   - If gold:
      - Allow unlimited returns
   - If silver:
      - Allow returns for purchases made in the last 6 months
   - If bronze:
      - Allow returns for purchases made in the last 90 days
   - If guest:
      - Allow returns for purchases made in the last 30 days
If customer wants to return an item due to wrong size:
   - Ask for customer's full name or account ID
   - Ask for username, email address, and order ID to validate purchase
   - Ask for membership level (guest, bronze, silver, gold)
   - If membership is guest:
      - If purchase was within last 30 days or customer has receipt/original packaging:
         - Allow return
      - Else:
         - Deny return
   - If membership is bronze:
      - If purchase was within last 90 days or customer has receipt/original packaging:
         - Allow return
      - Else:
         - Deny return
   - If membership is silver:
      - If purchase was within last 6 months or customer has receipt/original packaging:
         - Allow return
      - Else:
         - Deny return
   - If membership is gold:
      - Allow unlimited returns
   - If return is allowed:
      - Ask for customer's full address to generate shipping label
      - Ask how customer wants to process return (by mail, in store, drop off center)
      - Provide shipping label and return instructions

Table 8: Example of instructions in TOD-ProcBench.

Figure 3: User interface for collecting human annotation on the quality of the generated instructions.

**ABCD Conversation**

agent: hi. how can i help you?
customer: hello, i would like to check on the shipping status for my order as i have yet to receive it
agent: ok
agent: can you tell me your name, your account id, and the order id from the order?
customer: joseph banter, <account> , <order_id>
action: account has been pulled up for joseph banter.
action: identity verification in progress ...
agent: ok, what is your username and email address?
customer: <username> , <email>
action: purchase validation in progress ...
action: querying the system for an answer ...
agent: your order is still on track and on the way
customer: i have been waiting 9 days for it
customer: i think it might be lost, anyway i can get it resent?
agent: yes, it can be resent
agent: can you tell me your address
customer: <street_address> brighton, ny <zip_code>
action: order has been updated with <street_address> brighton, ny <zip_code>.
agent: ok, what is the brand and the type of product?
customer: michael kors jeans
action: a purchase of michael kors jeans was made.
agent: ok, the jeans have been resent
customer: thank you

**ABCD Workflow**

pull-up-account, verify-identity, validate-purchase, ask-the-oracle, update-order, make-purchase

**Examples of Task1 inputs**

**1. Partial Conversation:**
agent: hi. how can i help you?
customer: hello, i would like to check on the shipping status for my order as i have yet to receive it
agent: ok
agent: can you tell me your name, your account id, and the order id from the order?
customer: joseph banter, <account> , <order_id>
agent: ok, what is your username and email address?
customer: <username> , <email>
**1. Next Action:** validate-purchase

**2. Partial Conversation:**
agent: hi. how can i help you?
customer: hello, i would like to check on the shipping status for my order as i have yet to receive it
agent: ok
agent: can you tell me your name, your account id, and the order id from the order?
customer: joseph banter, <account> , <order_id>
agent: ok, what is your username and email address?
customer: <username> , <email>
agent: your order is still on track and on the way
customer: i have been waiting 9 days for it
customer: i think it might be lost, anyway i can get it resent?
agent: yes, it can be resent
agent: can you tell me your address
customer: < street_address> brighton, ny <zip_code>
agent: ok, what is the brand and the type of product?
customer: michael kors jeans
**2. Next Action:** make-purchase

Table 9: Examples of Task 1 benchmark.

---

**Fine-Grained Parameter Mismatch**

---

**Conversation**

"agent: hello. thanks for contacting the customer satisfaction team, how may i help you today?"

"customer: hi i never got my stuff i ordered. can you check it for me?"

"agent: it'd be my pleasure to look into that for you. may i start by getting your full name?"

"customer: sanya afzal"

**Instructions**

If customer has not received their order:

    If it has been less than 7 days since order was placed:

        - Inform customer that orders can take up to 7 days to arrive

        - Advise customer to wait a few more days and contact again if order still has not arrived after 7 days

    If it has been 7 days or more since order was placed:

        - Apologize for the delay and Verify customer's address and Place a new order to be shipped to the customer

    If order was placed a very long time ago (e.g. months), offer to expedite or overnight the new shipment

**Compliant Response**

"agent: okay sanya, what item were you expecting and when were you expecting it to be delivered?"

**Instruction-Violating Response (*fine-grained parameter mismatch*)**

"agent: Orders can sometimes take up to **11** business days to be delivered after being shipped. Let me know if you still have not received your order after 11 days and I'd be happy to look into it further for you."

---

Table 10: Example of a instruction-violating response resulting from the proposed compliance manipulation approach with fine-grained parameter mismatch.

---

**Single-Condition Action Replacement**

---

**Conversation**

"agent: hi! how can i help you today?"

"customer: hi! i need to return this jacket. it is the wrong size"

"agent: no problem. can i have your full name, username, email address, and order id, please?"

"customer: alessandro phoenix"

"customer: <username> <username>@email.c <email> <order_id>"

"agent: and what is your membership level?"

"customer: gold"

"customer: sanya afzal"

**Instructions**

If customer wants to return an item due to wrong size:

    - Ask for customer's full name or account ID

    - Ask for username, email address, and order ID to validate purchase

    - Ask for membership level (guest, bronze, silver, gold)

    - If membership is gold:

        - Allow unlimited returns

    ...

    - If return is allowed:

        - Ask for customer's full address to generate shipping label

        - Ask how customer wants to process return (by mail, in store, drop off center) and - Provide shipping label

**Compliant Response**

agent: okay, great! what is your address? i'll need it for the shipping label.

**Instruction-Violating Response (*single-condition action replacement*)**

I see your membership level is gold. Unfortunately, our policy does not allow unlimited returns for gold members. Could you please provide more details on why you need to return this item? I will review the return request based on the return policy.

---

Table 11: Example of a instruction-violating response resulting from the proposed compliance manipulation approach with single-condition action replacement.

**Multi-Condition Action Replacement**

**Conversation**

"agent: hello, thanks for contacting acmebrands. how can i help today?"

"customer: hi, this is albert sanders, gold member"

"customer: i have a premium subscription with acme brands, but i'm having a bit of trouble finding my account details"

"agent: i see, are you needing help accessing your account?"

"customer: yes, well, really i just want to know when my annual fee is due. i can pay it today if there's anything outstanding"

"action: account has been pulled up for albert sanders."

**Instructions**

If customer wants to check subscription status:

    - Ask for customer's full name, account ID, and order ID

    - Verify customer's identity

    - Check subscription status in the system

    - If subscription is active:

        - Provide due date and amount due

    ...

**Compliant Response**

"agent: do you have your account id and order id albert?"

**Instruction-Violating Response (*multi-condition action replacement*)**

"agent: According to our records, your annual subscription fee of $ 50 is due on March 1st. Let me check the status of your subscription in our system to confirm the details."

Table 12: Example of a instruction-violating response resulting from the proposed compliance manipulation approach with multi-condition action replacement.

**Conversation**
"turn 0: agent: hello, thanks for contacting acmebrands. how can i help today?"
"turn 1: customer: hi, this is albert sanders, gold member"
"turn 2: customer: i have a premium subscription with acme brands, but i'm having a bit of trouble finding my account details"
"turn 3: agent: i see, are you needing help accessing your account?"
"turn 4: customer: yes, well, really i just want to know when my annual fee is due. i can pay it today if there's anything outstanding"
"turn 5: action: account has been pulled up for albert sanders."

**Instructions**
If customer wants to check subscription status:
    - Ask for customer's full name, account ID, and order ID
    - Verify customer's identity
    - Check subscription status in the system
    - If subscription is active:
        - Provide due date and amount due
        - If amount is due today or past due:
            - Offer to take payment with card on file or new card
            - Process payment if customer wants to pay
    - If subscription is inactive:
        - Provide amount due and due date to reactivate
        - Offer to take payment with card on file or new card
        - Process payment if customer wants to reactivate

**Ground-Truth Response**
"turn 6: agent: do you have your account id and order id albert?"

**Extracted Original Instruction Section**
If customer wants to check subscription status:
    - Ask for customer's full name, account ID, and order ID
    - Verify customer's identity

**Manipulated Instructions**
If customer wants to check subscription status:
    - Provide due date and amount due
    - Check subscription status in the system

**Instruction-Violating Response**
"turn 6: agent: According to our records, your annual subscription fee of $50 is due on March 1st. Let me check the status of your subscription in our system to confirm the details."

Table 13: Example 1 of conversations with manipulated instructions and instruction-violating agent responses.

**Conversation**

"turn 0: agent: hello, how may i help you?",
"turn 1: customer: i noticed my phone number was wrong and i need to fix it.",
"turn 2: agent: let me help you with that.",
"turn 3: agent: so you just want to change the phone number on your account is that correct?",
"turn 4: customer: that's correct.",
"turn 5: agent: could i have your full name or account id?",
"turn 6: customer: albert sanders",
"turn 7: action: account has been pulled up for albert sanders"

**Instructions**

If customer wants to change their phone number:
   - Get customer's full name or account ID
   - Pull up customer's account
   - Get current phone number listed on the account
   - Verify customer's identity by asking for 2-3 of the following:
      - Zip code
      - Pin number
      - Username
      - Password
      - Email address
      - Previous order ID
   - If identity is verified:
      - Get new phone number from customer
      - Update account with new phone number
   - If identity cannot be verified:
      - Politely inform customer you cannot update account without proper verification
      - Offer to try verifying again or suggest customer contact support another way

**Ground-Truth Response**

"turn 8: agent: could i have the current phone number on your account?"

**Extracted Original Instruction Section**

If customer wants to change their phone number:
   - Get customer's full name or account ID
   - Pull up customer's account
   - Get current phone number listed on the account

**Manipulated Instructions**

If customer wants to change their phone number:
   - Get customer's full name or account ID
   - Pull up customer's account
   - Do not get current phone number listed on the account

**Instruction-Violating Response**

"turn 6: agent: I've pulled up your account, Albert. What new phone number would you like me to update it with?"

Table 14: Example 2 of conversations with manipulated instructions and instruction-violating agent responses.

## B   Complex Instruction Formats in TOD-ProcBench

In this section, we present a symbolic and natural language example of each of the three complex instruction formats, $f_1$ (Figure 4), $f_2$ (Figure 5), $f_3$ (Figure 6), in TOD-ProcBench.

Let $C$ be the set of all conditions and $A$ be the set of all actions. For a nested structure with depth $d$, each flattened condition-action pair is defined as:

$$(C_n, \mathbf{a}_n) \in \{(c_1 \wedge ... \wedge c_k, \langle a_1, ..., a_m \rangle) | c_j \in C, a_\ell \in A\}$$

where $C_n = c_1 \wedge ... \wedge c_k$ represents a conjunction of composite conditions, and $\mathbf{a}_n = \langle a_1, ..., a_m \rangle$ represents a sequence of actions that must be executed when $C_n$ is satisfied.

For example, the nested structure in $f_1$ [escapeinside=(**)] If (*$c_1$*): (*$a_1$*) If (*$c_2$*): (*$a_3$*) (*$a_4$*) (*$a_2$*) If (*$c_3$*): (*$a_5$*)

is flattened in $f_2$ to:

$$\{(c_1, \langle a_1 \rangle), (c_1 \wedge c_2, \langle a_3, a_4 \rangle), (c_1, \langle a_2 \rangle), (c_3, \langle a_5 \rangle)\}$$

.

A flattened structure in $f_2$ $\{(c_1, \langle a_1, a_2 \rangle)\}$ is then transformed in $f_3$ to $\{$"$conditions$" : $[c_1]$, "$actions$" : $[a_1, a_2]\}$.

```
<What_to_do>
If customer received an email confirmation showing incorrect quantity
of items ordered:
    - Identify customer by asking for full name or account ID
    - Get order ID from customer
    - Verify customer's identity
    - Check system records to confirm actual quantity ordered
    - If email was incorrect:
        - Inform customer of the correct quantity
        ordered based on system records
        - If order has already shipped:
            - Check shipping status
            - If delivered, start return process for extra item(s)
            - If not delivered yet, inform customer to contact again after
                receiving to initiate return
        - Offer refund for extra item(s) charged
    - If email was correct:
        - Explain reason for extra item(s) if possible
        - Start return process if applicable
</What_to_do>
```

Figure 4: Example of a TOD-ProcBench instruction document in format $f_1$.

```
<What_to_do>
If customer received an email confirmation showing incorrect quantity
of items ordered:
     - Identify customer by asking for full name or account ID
     - Get order ID from customer
     - Verify customer's identity
     - Check system records to confirm actual quantity ordered

If customer received an email confirmation showing incorrect quantity
of items ordered
AND email was incorrect:
     - Inform customer of the correct quantity ordered based on
     system records

If customer received an email confirmation showing incorrect quantity
of items ordered
AND email was incorrect AND order has already shipped:
     - Check shipping status

If customer received an email confirmation showing incorrect quantity
of items ordered
AND email was incorrect AND order has already shipped AND delivered:
     - Start return process for extra item(s)

If customer received an email confirmation showing incorrect quantity
of items ordered
AND email was incorrect AND order has already shipped AND not delivered yet:
     - Inform customer to contact again after receiving to initiate return

If customer received an email confirmation showing incorrect quantity
of items ordered
AND email was incorrect:
     - Offer refund for extra item(s) charged

If customer received an email confirmation showing incorrect quantity
of items ordered
AND email was correct:
     - Explain reason for extra item(s) if possible
     - Start return process if applicable
</What_to_do>
```

Figure 5: Example of a TOD-ProcBench instruction document in format $f_2$.

```
<What_to_do>
[
  {
    "Conditions": [
      "customer received an email confirmation showing incorrect quantity of
       items ordered"
    ],
    "Actions": [
      "Identify customer by asking for full name or account ID",
      "Get order ID from customer",
      "Verify customer's identity",
      "Check system records to confirm actual quantity ordered"
    ]
  },
  {
    "Conditions": [
      "customer received an email confirmation showing incorrect quantity of
       items ordered",
      "email was incorrect"
    ],
    "Actions": [
      "Inform customer of the correct quantity ordered based on system records"
    ]
  },
  {
    "Conditions": [
      "customer received an email confirmation showing incorrect quantity of
       items ordered",
      "email was incorrect",
      "order has already shipped"
    ],
    "Actions": [
      "Check shipping status"
    ]
  },
  {
    "Conditions": [
      "customer received an email confirmation showing incorrect quantity of
       items ordered",
      "email was incorrect",
      "order has already shipped",
      "delivered"
    ],
    "Actions": [
      "Start return process for extra item(s)"
    ]
  },
  {
    "Conditions": [
      "customer received an email confirmation showing incorrect quantity of
       items ordered",
      "email was incorrect",
      "order has already shipped",
      "not delivered yet"
    ],
    "Actions": [
      "Inform customer to contact again after receiving to initiate return"
    ]
  },
  {
    "Conditions": [
      "customer received an email confirmation showing incorrect quantity of
       items ordered",
      "email was incorrect"
    ],
    "Actions": [
      "Offer refund for extra item(s) charged"
    ]
  },
  {
    "Conditions": [
      "customer received an email confirmation showing incorrect quantity of
       items ordered",
      "email was correct"
    ],
    "Actions": [
      "Explain reason for extra item(s) if possible",
      "Start return process if applicable"
    ]
  }
]
</What_to_do>
```

Figure 6: Example of a TOD-ProcBench instruction document in format $f_3$.

# C    LLM Prompts

In this section, we present key prompts used for creating TOD-ProcBench and evaluating LLMs for each task.

---

**Instruction Generation LLM System Prompt (Part 1)**

---

**Conversation**

You are given a set of conversations between a customer and an agent. In all of the conversations the customer has a specific intent from the interaction with the agent. You are asked to read all the conversations in <conversations> tags for the specified intent in <intent> tag and generate a policy that the agent has followed to complete all the conversations in <conversations> tag.
hint: your generated policy should contain only one section in <what_to_do> tags and one <what_to_say> tags.

Here is a template for the policy that you are asked to generate. Each template has to have two sections:
1. What to do in <What_to_do> tags which shows some conditional statements and the steps to accomplish the task. You can specify each condition with 'if' and then have the condition and what needs to be done for that condition. Try to provide all details as an example directly mention different membership levels and the return dates for each level accordingly.
2. What to say in <What_to_say> tags which shows what are the important utterances that the agent has to say to accomplish all the defined steps in the what to do part.

Table 15: LLM system prompt (Part 1) to generate an instruction document given the set of conversations with similar user intents.

{continued from Table 15}

<example>

Here is an example of a policy for the following conversation in <example_conv> tag:

<example_conv>

"agent: hello, how are you today? how can i help you?",

"customer: i did a return but i don't want to get my refund as a gift card.",

"agent: can i please get your full name and account id?",

"customer: my name is [name1] but i'm not sure about the account id",

"agent: you are looking to get check instead of gift card?",

"customer: yes",

"action: account has been pulled up for [name1]",

"agent: what is your membership level",

"customer: gold",

"action: membership level of has been noted.",

"agent: since your are a gold member we can refund you via the check. Do you want to use your existing bank account?",

"customer: yes, thanks",

"action": changes have been entered.

"agent: no problem! is there anything else i can help you with, sanya?",

"customer: that will be all",

"agent: perfect. you have a nice day. goodbye!"

</example_conv>

Policy for this conversation is:

<What_to_do>

If customer wants to get check instead of gift card:

-If membership level is Gold or Silver:

    - Allow the request

    - Ask for the account info if you do not have

    - Process the request

-If membership level is guest or bronze:

    - if the refund has been initiated in 30 days:

        - Allow the request and ask for the account info and process it

    - if not:

        - Inform customer request cannot be processed

</What_to_do>

<What_to_say>

- Ask for customer's full name or account ID

- Check membership level

- Check refund initiated date based on membership level

- If the changing request is eligible:

    -Ask customer's bank account info if it does not exist.

    -Ask if customer wants to use existing bank account it if exists.

-If the request is not eligible:

    - Politely inform customer request cannot be processed and explain reasoning

    - Apologize for inconvenience

    - Offer to help with anything else

</What_to_say>

</example>

Table 16: LLM system prompt (Part 2) to generate an instruction document given the set of conversations with similar user intents.

---

**Instruction Generation LLM User Prompt**

---

**Conversation**
Here is the intent and all the conversations. Read all of them carefully and generate the policy accordingly.

<intent> *intent* </intent>

<conversations>

*convs*

</conversations>

The best policy for the given conversations is:

---

Table 17: LLM user prompt to generate an instruction document given the set of conversations with similar user intents.

---

**Complex Instruction Format Conversion LLM Prompt**

---

Convert nested conditional statements into two formats which should represent the same meaning in different formats:
1. A flattened text format where each line starts with a condition and lists actions to take
2. A flattened JSON format with an array of objects, each containing "conditions" and "actions" arrays
GENERAL INSTRUCTIONS:
- Identify all decision points and actions in the nested structure
- The entries for conditions and actions in each format must use the same natural language from the original input.
- Maintain all the original logic and relationships between conditions and actions
- Consecutive entries in the flattened formats should NOT have the exact same list for conditions. This effectively means that all condition-action pairs that are neighbors must have different lists for conditions. If the same list is used (same entries in the list or same empty list which specifically denotes all cases) for conditions in pairs that are consecutive, then combine all such consecutive pairs to only have one unique list of conditions and all corresponding actions in the same original sequence.
- The expectation is that a user of these flattened instructions will read them in order from top-down, first to last, evaluating each condition and if true executing the associated action(s). As such, your flattening must be recursive and preserve the same order of actions executed if the user were to follow the original SOP.
- Process each section separately if multiple sections are provided
INSTRUCTIONS for flattened text format:
- Each line should start with "If [condition]:" followed by indented actions
- When a condition depends on previous conditions, use "AND" to combine them (e.g., "If condition1 AND condition2:")
- For actions that apply in all cases, use "In all cases:" as the condition
INSTRUCTIONS for flattened json format:
- Represent each condition-action pair as "conditions": [...], "actions": [...]
- For actions that apply in all cases, use an empty list [] as the condition
You might find it easier to generate the flattened text format first and then generate the equivalent flattened json format. The motivation here is to flatten the nested hierarchy of conditional statements so that one can simply go down the list from the first line in a flattened format to the last line in a flattened format and only execute actions corresponding to conditional statements that are true.
Please convert this into both flattened formats and provide your output.
Input:
***json_data***
Provide a single dictionary so that json.loads() can parse your perfectly formatted response. Your final response should be a json with the following keys (note that the first 3 keys are the same as original input and the values should be preserved as well):
"intent"  "what_to_do",  "what_to_say",  "flattened_text_what_to_do",  "flattened_text_what_to_say", "json_what_to_do", "json_what_to_say",

---

Table 18: LLM prompt to convert complex instruction format from $f_1$ to $f_2$ and $f_3$.

---

**Multilingual Conversation Translation LLM Prompt**

---

Translate this conversation from english to ***target_language***. For both the agent and customer, use the native conversation style. However, the tone and formality should match the tone and formality for each role respectively in the original English conversation. This means, for the agent, the tone should be professional and the formality should be formal, even if the customer seems to be speaking informally. respect the masculine and feminine assignments of vocabulary in the target language. The output format should be the same as the input conversation: A string with turns separated by \n exactly as done in the input transcript of the conversation, representing either the "agent" or "customer" speaking, with the actual dialogue text in the target language. Avoid using English except for the role names as explained below or placeholders indicated in the conversation. The roles in the conversation list should still be represented as "agent" or "customer". Only respond with this conversation transcript string and nothing else.

Original english conversation transcript: ***conversation***

Translated conversation transcript:

---

Table 19: LLM prompt to translate conversations in TOD-ProcBench.

---

**Multilingual Conversation Translations LLM Judge Prompt**

---

Evaluate the translation quality of the translated ***language*** conversation with respect to the original English conversation and answer three questions with Yes or No:

English conversation: ***english_conversation***

Translated conversation: ***translated_conversation***

Please only provide your answers as three comma-separated Yes/No responses to the following questions and do not provide anything else in your answer:

1. Overall, did the translated conversation preserve the same meaning in each agent/customer turn as the original English conversation?

2. Overall, did the sentiment, tone, and formality of each agent/customer turn in the translated conversation match the respective agent/customer turn in the original English conversation?

3. Except for placeholder text or technical terms which do not have proper translations to the target language, is there any use of English in the translated text?

---

Table 20: LLM-as-a-Judge prompt to evaluate the quality of translated conversations in TOD-ProcBench.

---

**LLM System Prompt for Task 1**

---

You are given a part of conversation between a customer and an agent in <conversation> tag. In addition, you are given a set of policies in <policies> tag including "what_to_do" and "what_to_say" that are instructed to an agent to follow in its interaction with the customer to solve customer's provided issue.

You are also given a list of actions in <actions> that shows agent's next action to conduct in each point of the conversation.

Each line in <actions> shows one possible agent's action and the conditions necessary to select that action. The conditions are based on what customer has asked and looked for in the last turns of the conversation. In order to select the agent's next action you need to pay attention to the last agent's and customer's turns from the conversation in <conversation> tags.

Your task is as following:
get the two most relevant policies:

    1) read all the policies carefully

    2) for the given conversation select the two most matching and relevant policies based on the customer's intent from the interaction. Your selected policies should meet both of the following conditions.

        a) policies state has user's provided intent or problem as an "if" condition.

        b) policies include the steps that agent has followed in its interaction in the conversation to solve the problem.

get agent's next action:

    1)read all the list of actions in <actions> and the conditions necessary to select each action.

    2)read only the last agent and customer turns from <conversation>

    3)find the most similar action from <actions> tag by comparing customer's last turn from conversation and each action's condiditon from <actions>.

Look into following examples for identifying next agent's action.

<examples>

<example>

<conversation>

"agent: hi! how can i help you?",

"customer: i'd like to know why i was charged twice for my subscription.",

"agent: sure, sorry about that, let me take a look",

"agent: may i have your name please?",

"customer: joseph banter",

"agent: thanks, and your account id and order id please?",

"customer: account id: <account_id>",

"customer: order id: <order_id>",

"agent: thanks! do you know your membership level?",

"customer: guest",

<conversation>

<reason> To get the agent's next action we pay attention to the last agent's and customer's turns in the <conversation> which is "thanks! do you know your membership level?" and "guest" where the agent has asked customer about her membership status and customer has provided it. It is exactly similar to what triggers "membership" action in <actions> tag that is "agent has asked customer about her membership status and customer has provided her membership level". Therefore the agent's next action is "membership".

<reason>

<example>

...

</examples>

Start your generation first by providing a reasoning of why you have selected the agent's next action (similar to provided examples) and the two most relevant policies in <reason> tags. Then return one <output> tag that includes THREE piece of information: the most relevant next action from <actions> and two most relevant policies from <policies> all separated by ','. The format of the output should be your provided reason in <reason> tags and then <output>a,n,m</output> without any nextline character where a is the most relevant next action from <actions> tag that agent should do given the last agent and customer turns in the conversation, n and m are the most relevant policies to the conversation.

---

Table 21: LLM system prompt for Task1.

---

**LLM User Prompt for Task 1**

---

List of policies: <policies> *policies* </policies>
List of next actions:
<actions>
**search-policy**: in the last customer turn from given <conversation> customer has asked policy related questions but agent has not searched the policy and provided info yet.
**make-password**: in the last customer turn from given <conversation> customer has forgotten her password and customer answered to the agent's security questions to make new password but agent has not made the password yet.
**enter-details**: in the last customer turn from given <conversation> customer has provided personal information such as pin, address, email, order id and etc that agent has asked. Agent next will enter those details or agent directly mentions that is going to note those details.
**membership**: in the last customer turn from given <conversation> customer has provided her membership status since the agent has asked about.
**notify-team**: in the last turn from given <conversation> agent has collected customer's issue and promised customer that is going to notify someone regarding that issue.
**make-purchase**: in the last customer turn from given <conversation> customer has requested agent to create a new order.
**search-pricing**: in the last customer turn from given <conversation> customer has requested to know about price of something and agent has not provided the price yet.
**record-reason**: in the last customer turn from given <conversation> customer has complained about some issue and mentioned the reason of that issue.
**search-shirt**: in the last customer turn from given <conversation> customer had question relevant to shirt and agent has not searched about shirt yet.
**search-jacket**: in the last customer turn from given <conversation> customer had question relevant to jacket and agent has not searched about jacket yet.
**search-jeans**: in the last customer turn from given <conversation> customer had question relevant to jeans and agent has not searched about jeans yet.
**search-boots**: in the last customer turn from given <conversation> customer had question relevant to boots and agent has not searched about boots yet.
**search-timing**: in the last customer turn from given <conversation> customer had question relevant to timing such as hours, due date and questions with 'when' and agent has not searched it to get answer yet.
**promo-code**: in the last customer turn from given <conversation> customer wanted to get a new promo code or agent has suggested to create promo-code to keep the customer happy and customer agreed. This action in only regarding creating new promo-code not other promo-code related requests.
**log-out-in**: in the last customer turn from given <conversation> customer has told about some issue and agent still has not provided logging-in and out as a solution.
**shipping-status**: in the last customer turn from given <conversation> customer has provided shipping status of the order to the agent.
**offer-refund**: in the last customer turn from given <conversation> customer has provided information such as the item brand, its cost to get a refund from agent since agent has suggested to give refund to the customer.
**validate-purchase**: in the last customer turn from given <conversation> customer has provided some order relevant information such as order id, account id to retrieve and validate whether there was such an order.
**instructions**: in the last customer turn from given <conversation> customer has talked about some issue but agent has not yet looked for some instruction and solutions such as how clearing cookies on browser.

{continued in Table }

---

Table 22: LLM user prompt (Part 1) for Task1.

**search-membership**: in the last customer turn from given <conversation> customer has asked questions regarding different memberships and agent has not explained those search and has not provided information about membership levels.

**select-faq**: in the last customer turn from given <conversation> customer has asked questions and agent has not done the search yet and also has not selected faq pages yet to answer those questions.

**try-again**: in the last customer turn from given <conversation> customer has talked about some issue and agent has suggested to recheck and redo some actions to make sure the issue still persists.

**pull-up-account**: in the last customer turn from given <conversation> customer has provided her personal information such as name, and agent has not used it yet to pull up customer's account.

**subscription-status**: in the last customer turn from given <conversation> customer has provided her account but agent has not looked for the customer's subscription status yet.

**search-faq**: in the last customer turn from given <conversation> customer has asked a question and the agent has not searched and has not looked into faq to answer the question.

**send-link**: in the last customer turn from given <conversation> customer has provided her email such that agent can send a link to provide further information or as a receipt but agent has not sent the link yet.

**update-account**: in the last customer turn from given <conversation> customer has asked for changing some information such as credit card, subscription status, ... and agent has not updated them yet or agent has not applied some changes such as extension yet.

**update-order**: in the last customer turn from given <conversation> customer has provided new information regarding the order and agent has not used them to change and update item, its address and etc yet.

**ask-the-oracle**: in the last customer turn from given <conversation> customer has provided some claim and agent needs to check the system and look into the validity of customer's provided claim and order/return/refund details. ask-the-oracle usually happens when customer has already provided all her personal information and agent completely knows about the customer and her situation.

**verify-identity**: in the last customer turn from given <conversation> customer has provided her identity by telling the account id, email, phone and etc. Right after providing identity information agent should conduct verify-identity action.

**empty**: in the last customer turn from given <conversation> customer has no further requests or the agent's next action does not match with any of the previous actions or the customer has already achieved her goal meaning that agent does not have next action.

</actions>

<conversation> *conv* </conversation>

Table 23: LLM user prompt (Part 2) for Task1.

**LLM System Prompt for Task 2 Approach 1**

You are given a part of conversation between a customer and an agent in `<conversation>` tags, where customer interacts with the agent for a specific intent. In addition, you are given the policy in `<policy>` tags that agent needs to follow in order to help the customer meet her intent alongside the agent's next utterance in `<response>` tag.

Your task is to identify whether the agent's next response has followed the policy or not in other word whether the agent's next response is compliant with the policy or not. In order to assess the compliance of the agent's next response, you need to specify according to the given conversation which part of the policy is the conversation right now and according to the policy what is the next best action to choose from. The agent's next response should be compliant with that action from the policy and provides next steps in a correct order following the policy. If the next action provided in the agent's next response is not exactly the next step coming from the policy label that response as not compliant.

Look into following examples:

Examples

Example 1

```
<conversation>
turn 0: agent: hi! thanks for contacting acme support. how can i help you today?
turn 1: customer: starting to wonder if my package might not be lost
turn 2: agent: sorry to hear that.
turn 3: customer: been 5 days since i've ordered and haven't gotten it
turn 4: agent: let's see what we can find out.
turn 5: agent: may i have your full name?
turn 6: customer: joseph banter
turn 7: customer: the order in question were calvin klein boots <order_id>
turn 8: agent: one moment mr. banter.
turn 9: agent: okay, can i get your username and email address?
turn 10: customer: <username>
turn 11: customer: <email>
turn 12: agent: got it, thanks.
turn 13: agent: hmm, and you say you've been waiting 5 days for your item?
turn 14: customer: that's right
turn 15: customer: dunno if it's time to consider it lost or what
</conversation>

<response>
turn 16: agent: typically, we wait three weeks until we take that type of action.
</response>

<policy>
- Ask for customer's full name, username, email address, and order ID to validate the purchase
- Check how many days the customer has been waiting for the order
- If less than 7 days, explain that orders can sometimes take up to 7 days and ask for patience
- If 7 days or more, apologize for the delay and let them know you will resend the new order
    - Ask for their current mailing address to send the new order
    - Confirm the brand, product type, and any other details about the missing item
    - Let them know a replacement order has been placed and provide an estimate on when it will arrive
- Offer further assistance or ask if there is anything else needed
- Thank the customer for their patience and understanding
</policy>

<reason>
since the <response> response has provided three weeks as a deadline to wait to receive the order
while in the policy it is 7 days therefore the response is not compliant with the policy.
</reason>

<compliant>
no
</compliant>
```

Instructions

First provide reasoning of how did you assess the compliance of the agent's next response to the policy in `<reason>` tag and then generate the output in `<compliant>` tag. Generate "yes" as the output if the next agent's response is compliant with policy and "no" otherwise.

Table 24: LLM system prompt for Task 2 Approach 1.

## LLM System Prompt for Task 2 Approach 2

You are given a part of conversation between a customer and an agent in `<conversation>` tags, where customer interacts with the agent for a specific intent. In addition, you are given the policy in `<policy>` tags that agent needs to follow in order to help the customer meet her intent alongside the agent's next utterance in `<response>` tag.

Your task is to identify whether the agent's next response has followed the policy or not in other word whether the agent's next response is compliant with the policy or not. In order to assess the compliance of the agent's next response, you need to first generate the most relevant policy to the given conversation + response and then compare whether the generated policy is some part of the overall given policy or not.

Look into following examples: Example 1

```
<conversation>
turn 0: agent: hi! thanks for contacting acme support. how can i help you today?
turn 1: customer: starting to wonder if my package might not be lost
turn 2: agent: sorry to hear that.
turn 3: customer: been 5 days since i've ordered and haven't gotten it
turn 4: agent: let's see what we can find out.
turn 5: customer: may i have your full name?
turn 6: customer: joseph banter
turn 7: customer: the order in question were calvin klein boots <order_id>
turn 8: agent: one moment mr. banter.
turn 9: agent: okay, can i get your username and email address?
turn 10: customer: <username>
turn 11: customer: <email>
turn 12: agent: got it, thanks.
turn 13: agent: hmm, and you say you've been waiting 5 days for your item?
turn 14: customer: that's right
turn 15: customer: dunno if it's time to consider it lost or what
</conversation>

<response>
turn 16: agent: typically, we wait three weeks until we take that type of action.
</response>

<policy>
- Ask for customer's full name, username, email address, and order ID to validate the purchase
- Check how many days the customer has been waiting for the order
- If less than 7 days, explain that orders can sometimes take up to 7 days and ask for patience
- If 7 days or more, apologize for the delay and let them know you will resend the order
    - Ask for their current mailing address to send the new order
    - Confirm the brand, product type, and any other details about the missing item
    - Let them know a replacement order has been placed and provide an estimate on when it will arrive
- Offer further assistance or ask if there is anything else needed
- Thank the customer for their patience and understanding
</policy>

<reason>
First I generate the following policy given the conversation and the response:
<gen_policy>
- Check how many days the customer has been waiting for the order
- If less than three weeks, explain that orders can sometimes take up to three weeks and ask for patience
</gen_policy>
Since the generated policy does not exist in the given policy
therefore I specify the response as "no" complining with the policy.
</reason>

<compliant>
no
</compliant>
```

Instructions:
First provide reasoning of how did you assess the compliance of the agent's next response to the policy in `<reason>` tag. In `<reason>` tag you should generate the policy given the conversation and response in `<gen_policy>` tags and then generate the output in `<compliant>` tag. Generate "yes" as the output if the `<gen_policy>` is part of `<policy>` and "no" otherwise.

Table 25: LLM system prompt for Task 2 Approach 2.

## LLM User Prompt for Task 2

<conversation> ***conversation*** </conversation>
<response> ***ground_truth_response*** </response>
<policy> ***policy*** </policy>

Table 26: LLM user prompt for Task 2.

**LLM System Prompt for Task 3**

You are given a part of conversation between a customer and an agent in <conversation> tags, where customer interacts with the agent for a specific intent. In addition, you are given the policy in <policy> tags that agent needs to follow in order to help the customer meet her intent.

Your task is to identify the steps that the agent has followed up to the current point from the policy and according to the next step in the policy generate an answer which is compliant with the policy.

To generate the agent's next response do the following steps:

1. read the whole conversation in <conversation> tag.

2. read the whole policy in <policy> tag.

3. based on the conversation, identify all the steps that agent has followed from the policy and get the next action from the policy to conduct.

4. generate the agent's next response, using the same language as the conversation transcript, based on the retrieved next step from the policy such that:

a) all the provided information should be covered in the policy's next step.

b) do NOT hallucinate and do NOT generate detailed information that do not exist in the provided policy's next step.

c) generate short responses that directly address the next best action coming from the policy.

First provide reasoning of how did you generate the next response in <reason> tag and then generate the next agent's response, using the same language as the conversation transcript, in <response> tags for the given conversation that matches with one next step in the provided policy. The generated response has to be short (less than three sentences).

Table 27: LLM system prompt for Task 3.

**LLM User Prompt for Task 3**

<conversation> ***conversation*** </conversation>
<policy> ***policy*** </policy>

Table 28: LLM user prompt for Task 3.

# D Results

In this section, we present the results tables for experiments which were not presented in tables within the main text due to space constraints.

| Model | Language | Overall | Instruction Retrieval | Begin | Middle | End |
|---|---|---|---|---|---|---|
| Claude3.7-Sonnet | FR | 0.4041 | 0.8286 | 0.4141 | 0.3908 | 0.4065 |
| Claude3.7-Sonnet | ES | 0.4005 | 0.8329 | 0.3955 | 0.4100 | 0.3963 |
| Claude3.7-Sonnet | EN | 0.3932 | 0.8205 | 0.3959 | 0.3895 | 0.3939 |
| Claude3.7-Sonnet | DE | 0.3896 | 0.8195 | 0.3885 | 0.3926 | 0.3877 |
| Claude3.7-Sonnet | HI | 0.3882 | 0.8137 | 0.3785 | 0.4004 | 0.3867 |
| Claude3.7-Sonnet | AR | 0.3778 | 0.8053 | 0.3820 | 0.3734 | 0.3775 |
| Claude3.7-Sonnet | ZH | 0.3768 | 0.8096 | 0.3808 | 0.3730 | 0.3761 |
| Claude3.5-Sonnet-V2 | FR | 0.3613 | 0.7372 | 0.3615 | 0.3695 | 0.3520 |
| Claude3.5-Sonnet-V2 | DE | 0.3524 | 0.7276 | 0.3483 | 0.3582 | 0.3510 |
| Claude3.5-Sonnet-V2 | ES | 0.3472 | 0.7227 | 0.3518 | 0.3521 | 0.3361 |
| Claude3.5-Sonnet-V1 | HI | 0.3365 | 0.7260 | 0.3386 | 0.3399 | 0.3303 |
| Claude3.5-Sonnet-V2 | EN | 0.3312 | 0.7311 | 0.3270 | 0.3333 | 0.3341 |
| Claude3.5-Sonnet-V1 | FR | 0.3299 | 0.7112 | 0.3522 | 0.3190 | 0.3144 |
| Claude3.5-Sonnet-V1 | DE | 0.3283 | 0.7088 | 0.3386 | 0.3181 | 0.3269 |
| Claude3.5-Sonnet-V1 | ES | 0.3268 | 0.7073 | 0.3282 | 0.3316 | 0.3197 |
| Claude3.5-Sonnet-V1 | ZH | 0.3216 | 0.7103 | 0.3185 | 0.3359 | 0.3095 |
| Claude3.5-Sonnet-V2 | ZH | 0.3209 | 0.6977 | 0.3212 | 0.3129 | 0.3293 |
| Claude3.5-Sonnet-V1 | AR | 0.3206 | 0.7063 | 0.3293 | 0.3146 | 0.3163 |
| Claude3.5-Sonnet-V2 | HI | 0.3141 | 0.6842 | 0.3317 | 0.3102 | 0.2965 |
| Qwen3-14B | EN | 0.3122 | 0.7792 | 0.1645 | 0.2828 | 0.5289 |
| Qwen3-14B | HI | 0.3017 | 0.7582 | 0.1718 | 0.2693 | 0.4995 |
| Qwen3-14B | ZH | 0.2983 | 0.7392 | 0.1521 | 0.2654 | 0.5169 |
| Qwen3-14B | FR | 0.2976 | 0.7444 | 0.1540 | 0.2688 | 0.5082 |
| Claude3.5-Sonnet-V2 | AR | 0.3036 | 0.6764 | 0.3007 | 0.3102 | 0.2999 |
| Claude3.5-Sonnet-V1 | EN | 0.2963 | 0.6491 | 0.3019 | 0.2911 | 0.2951 |
| Qwen3-14B | AR | 0.2934 | 0.7135 | 0.1726 | 0.2505 | 0.4913 |
| Qwen3-14B | ES | 0.2874 | 0.7479 | 0.1521 | 0.2519 | 0.4952 |
| Qwen3-14B | DE | 0.2786 | 0.7430 | 0.1242 | 0.2558 | 0.4961 |
| Gemma3-27B-IT | EN | 0.2221 | 0.6138 | 0.0770 | 0.2144 | 0.4113 |
| Gemma3-27B-IT | DE | 0.2088 | 0.5655 | 0.0708 | 0.2044 | 0.3857 |
| Gemma3-27B-IT | ES | 0.1978 | 0.5646 | 0.0658 | 0.1834 | 0.3780 |
| Gemma3-27B-IT | ZH | 0.1921 | 0.5324 | 0.0550 | 0.1891 | 0.3664 |
| Gemma3-27B-IT | HI | 0.1878 | 0.5355 | 0.0701 | 0.1786 | 0.3447 |
| Gemma3-27B-IT | FR | 0.1877 | 0.5175 | 0.0441 | 0.1861 | 0.3684 |
| Gemma3-27B-IT | AR | 0.1585 | 0.4615 | 0.0433 | 0.1464 | 0.3153 |
| Llama3.3-70B | EN | 0.1358 | 0.2042 | 0.1374 | 0.1338 | 0.1360 |
| Llama3.3-70B | ES | 0.1280 | 0.1976 | 0.1327 | 0.1298 | 0.1201 |
| Llama3.3-70B | FR | 0.1146 | 0.1770 | 0.1188 | 0.1181 | 0.1056 |
| Llama3.3-70B | DE | 0.1126 | 0.1729 | 0.1262 | 0.1089 | 0.0998 |
| Llama3.3-70B | ZH | 0.0840 | 0.1688 | 0.0851 | 0.0780 | 0.0892 |
| Llama3.3-70B | AR | 0.0299 | 0.0653 | 0.0298 | 0.0270 | 0.0333 |
| Llama3.3-70B | HI | 0.0240 | 0.0631 | 0.0252 | 0.0227 | 0.0241 |

Table 29: Task 1 accuracy of top $k = 1$ instruction retrieval and next agent action prediction (only format $f_1$). Results are shown across different models and conversation languages (Arabic (AR), Chinese (ZH), English (EN), French (FR), German (DE), Hindi (HI), Spanish (ES)).

| Model | Language | Overall | Instruction Retrieval | Begin | Middle | End |
|---|---|---|---|---|---|---|
| Claude3.7-Sonnet | EN | 0.4310 | 0.8943 | 0.4481 | 0.4074 | 0.4359 |
| Claude3.7-Sonnet | FR | 0.4287 | 0.9022 | 0.4346 | 0.4283 | 0.4219 |
| Claude3.7-Sonnet | ES | 0.4246 | 0.9055 | 0.4272 | 0.4179 | 0.4286 |
| Claude3.5-Sonnet-V2 | DE | 0.4237 | 0.8802 | 0.4276 | 0.4222 | 0.4204 |
| Claude3.7-Sonnet | AR | 0.4234 | 0.8998 | 0.4149 | 0.4314 | 0.4253 |
| Claude3.7-Sonnet | DE | 0.4205 | 0.8992 | 0.4327 | 0.4126 | 0.4142 |
| Claude3.7-Sonnet | HI | 0.4198 | 0.9003 | 0.4129 | 0.4227 | 0.4253 |
| Claude3.5-Sonnet-V2 | FR | 0.4194 | 0.8838 | 0.4319 | 0.4200 | 0.4031 |
| Claude3.7-Sonnet | ZH | 0.4146 | 0.8954 | 0.4122 | 0.4070 | 0.4262 |
| Claude3.5-Sonnet-V2 | ES | 0.4146 | 0.8825 | 0.4187 | 0.4109 | 0.4137 |
| Claude3.5-Sonnet-V2 | HI | 0.4040 | 0.8819 | 0.3913 | 0.4131 | 0.4098 |
| Claude3.5-Sonnet-V2 | EN | 0.3941 | 0.8721 | 0.4036 | 0.3904 | 0.3862 |
| Claude3.5-Sonnet-V2 | AR | 0.3929 | 0.8753 | 0.3924 | 0.4048 | 0.3804 |
| Claude3.5-Sonnet-V2 | ZH | 0.3922 | 0.8750 | 0.3901 | 0.3983 | 0.3881 |
| Claude3.5-Sonnet-V1 | FR | 0.3745 | 0.8159 | 0.3808 | 0.3721 | 0.3693 |
| Claude3.5-Sonnet-V1 | DE | 0.3744 | 0.8207 | 0.3750 | 0.3725 | 0.3756 |
| Claude3.5-Sonnet-V1 | HI | 0.3706 | 0.8179 | 0.3769 | 0.3625 | 0.3717 |
| Claude3.5-Sonnet-V1 | ES | 0.3689 | 0.8202 | 0.3731 | 0.3756 | 0.3563 |
| Claude3.5-Sonnet-V1 | AR | 0.3532 | 0.8130 | 0.3595 | 0.3503 | 0.3486 |
| Claude3.5-Sonnet-V1 | ZH | 0.3491 | 0.8066 | 0.3518 | 0.3521 | 0.3423 |
| Claude3.5-Sonnet-V1 | EN | 0.3485 | 0.7830 | 0.3533 | 0.3490 | 0.3419 |
| Qwen3-14B | ES | 0.3340 | 0.8552 | 0.1788 | 0.2998 | 0.5651 |
| Qwen3-14B | AR | 0.3286 | 0.8557 | 0.1753 | 0.2824 | 0.5709 |
| Qwen3-14B | EN | 0.3279 | 0.8419 | 0.1668 | 0.2880 | 0.5728 |
| Qwen3-14B | DE | 0.3273 | 0.8441 | 0.1548 | 0.3041 | 0.5680 |
| Qwen3-14B | HI | 0.3207 | 0.8376 | 0.1560 | 0.3015 | 0.5473 |
| Qwen3-14B | ZH | 0.3151 | 0.8260 | 0.1513 | 0.2932 | 0.5434 |
| Qwen3-14B | FR | 0.3122 | 0.8199 | 0.1389 | 0.2906 | 0.5521 |
| Gemma3-27B-IT | HI | 0.2540 | 0.7230 | 0.0786 | 0.2558 | 0.4706 |
| Gemma3-27B-IT | EN | 0.2528 | 0.7283 | 0.0809 | 0.2414 | 0.4797 |
| Llama3.3-70B | EN | 0.2497 | 0.6612 | 0.2620 | 0.2366 | 0.2488 |
| Gemma3-27B-IT | FR | 0.2444 | 0.6970 | 0.0592 | 0.2227 | 0.4990 |
| Gemma3-27B-IT | DE | 0.2408 | 0.6961 | 0.0728 | 0.2248 | 0.4677 |
| Gemma3-27B-IT | AR | 0.2364 | 0.7102 | 0.0813 | 0.2209 | 0.4470 |
| Llama3.3-70B | FR | 0.2353 | 0.6872 | 0.2512 | 0.2248 | 0.2271 |
| Gemma3-27B-IT | ES | 0.2349 | 0.6885 | 0.0677 | 0.2044 | 0.4769 |
| Llama3.3-70B | DE | 0.2307 | 0.6578 | 0.2194 | 0.2423 | 0.2319 |
| Llama3.3-70B | ES | 0.2305 | 0.6859 | 0.2357 | 0.2322 | 0.2223 |
| Gemma3-27B-IT | ZH | 0.2262 | 0.6837 | 0.0565 | 0.2266 | 0.4373 |
| Llama3.3-70B | ZH | 0.1293 | 0.4207 | 0.1277 | 0.1216 | 0.1398 |
| Llama3.3-70B | AR | 0.0511 | 0.1906 | 0.0534 | 0.0449 | 0.0550 |
| Llama3.3-70B | HI | 0.0390 | 0.1769 | 0.0430 | 0.0375 | 0.0357 |

Table 30: Task 1 accuracy of top $k = 2$ instructions retrieval and next agent action prediction (only format $f_1$). Results are shown across different models and conversation languages (Arabic (AR), Chinese (ZH), English (EN), French (FR), German (DE), Hindi (HI), Spanish (ES)).

| Model | k | Data Format | Overall | Instruction Retrieval | Begin | Middle | End |
|---|---|---|---|---|---|---|---|
| Claude3.7-Sonnet | 2 | $f_1$ | 0.4310 | 0.8943 | 0.4481 | 0.4074 | 0.4359 |
| Claude3.7-Sonnet | 2 | $f_2$ | 0.4215 | 0.9071 | 0.4122 | 0.4366 | 0.4166 |
| Claude3.7-Sonnet | 2 | $f_3$ | 0.4205 | 0.8880 | 0.4191 | 0.4179 | 0.4253 |
| Claude3.5-Sonnet-V2 | 2 | $f_3$ | 0.4182 | 0.8765 | 0.4214 | 0.4126 | 0.4204 |
| Claude3.7-Sonnet | 1 | $f_2$ | 0.3948 | 0.8267 | 0.3878 | 0.3834 | 0.4161 |
| Claude3.5-Sonnet-V2 | 2 | $f_1$ | 0.3941 | 0.8721 | 0.4036 | 0.3904 | 0.3862 |
| Claude3.5-Sonnet-V2 | 2 | $f_2$ | 0.3939 | 0.8631 | 0.4005 | 0.3956 | 0.3838 |
| Claude3.7-Sonnet | 1 | $f_1$ | 0.3932 | 0.8205 | 0.3959 | 0.3895 | 0.3939 |
| Claude3.7-Sonnet | 1 | $f_3$ | 0.3869 | 0.8163 | 0.3932 | 0.3821 | 0.3843 |
| Claude3.5-Sonnet-V1 | 2 | $f_3$ | 0.3716 | 0.8304 | 0.3750 | 0.3730 | 0.3660 |
| Claude3.5-Sonnet-V1 | 2 | $f_3$ | 0.3485 | 0.7830 | 0.3533 | 0.3490 | 0.3419 |
| Claude3.5-Sonnet-V1 | 2 | $f_2$ | 0.3383 | 0.7910 | 0.3437 | 0.3259 | 0.3452 |
| Claude3.5-Sonnet-V2 | 1 | $f_1$ | 0.3312 | 0.7311 | 0.3270 | 0.3333 | 0.3341 |
| Qwen3-14B | 2 | $f_1$ | 0.3279 | 0.8419 | 0.1668 | 0.2880 | 0.5728 |
| Claude3.5-Sonnet-V1 | 1 | $f_3$ | 0.3187 | 0.7057 | 0.3185 | 0.3342 | 0.3018 |
| Qwen3-14B | 1 | $f_1$ | 0.3122 | 0.7792 | 0.1645 | 0.2828 | 0.5289 |
| Claude3.5-Sonnet-V1 | 1 | $f_1$ | 0.2963 | 0.6491 | 0.3019 | 0.2911 | 0.2951 |
| Qwen3-14B | 2 | $f_2$ | 0.2961 | 0.8418 | 0.1250 | 0.2623 | 0.5468 |
| Qwen3-14B | 1 | $f_2$ | 0.2845 | 0.7713 | 0.1142 | 0.2584 | 0.5256 |
| Qwen3-14B | 2 | $f_3$ | 0.2712 | 0.7879 | 0.1130 | 0.2248 | 0.5198 |
| Gemma3-27B-IT | 2 | $f_2$ | 0.2559 | 0.7559 | 0.0755 | 0.2392 | 0.4990 |
| Claude3.5-Sonnet-V2 | 1 | $f_3$ | 0.2543 | 0.5152 | 0.2604 | 0.2606 | 0.2396 |
| Gemma3-27B-IT | 2 | $f_1$ | 0.2528 | 0.7283 | 0.0809 | 0.2414 | 0.4797 |
| Llama3.3-70B | 2 | $f_1$ | 0.2497 | 0.6612 | 0.2620 | 0.2366 | 0.2488 |
| Gemma3-27B-IT | 2 | $f_3$ | 0.2362 | 0.6717 | 0.07314 | 0.2209 | 0.4561 |
| Llama3.3-70B | 2 | $f_2$ | 0.2362 | 0.5924 | 0.2318 | 0.2327 | 0.2454 |
| Gemma3-27B-IT | 1 | $f_1$ | 0.2221 | 0.6138 | 0.0770 | 0.2144 | 0.4113 |
| Qwen3-14B | 1 | $f_3$ | 0.2110 | 0.6368 | 0.0681 | 0.1887 | 0.4137 |
| Claude3.5-Sonnet-V2 | 1 | $f_2$ | 0.2084 | 0.4654 | 0.2105 | 0.2044 | 0.2102 |
| Llama3.3-70B | 2 | $f_3$ | 0.2065 | 0.5376 | 0.2020 | 0.2013 | 0.2179 |
| Claude3.5-Sonnet-V1 | 1 | $f_2$ | 0.2048 | 0.5038 | 0.2051 | 0.2026 | 0.2068 |
| Gemma3-27B-IT | 1 | $f_2$ | 0.1847 | 0.5225 | 0.0557 | 0.1699 | 0.3616 |
| Gemma3-27B-IT | 1 | $f_3$ | 0.1740 | 0.4960 | 0.0526 | 0.1695 | 0.3303 |
| Llama3.3-70B | 1 | $f_2$ | 0.1447 | 0.2503 | 0.1463 | 0.1477 | 0.1393 |
| Llama3.3-70B | 1 | $f_1$ | 0.1358 | 0.2042 | 0.1374 | 0.1338 | 0.1360 |
| Llama3.3-70B | 1 | $f_3$ | 0.1299 | 0.2444 | 0.1413 | 0.1268 | 0.1191 |

Table 31: Task 1 accuracy of top $k = 1$ and $k = 2$ instructions retrieval and next agent action prediction (only English conversations). Results are shown across different models and the three complex instruction formats ($f_1$, $f_2$, $f_3$).

| Model | Approach 1 | | | Approach 2 | | |
|---|---|---|---|---|---|---|
| | $f_1$ | $f_2$ | $f_3$ | $f_1$ | $f_2$ | $f_3$ |
| Claude3.7-Sonnet | **0.7608** | **0.7525** | **0.7528** | **0.7298** | **0.7346** | **0.7298** |
| Claude3.5-Sonnet-V2 | 0.7215 | 0.7147 | <u>0.7238</u> | <u>0.6874</u> | 0.6826 | 0.6847 |
| Claude3.5-Sonnet-V1 | 0.6847 | 0.6930 | <u>0.6945</u> | <u>0.6854</u> | 0.6771 | 0.6748 |
| Llama3.3-70B | <u>0.7283</u> | <u>0.7283</u> | 0.7159 | 0.5636 | <u>0.5747</u> | 0.5368 |
| Qwen3-14B | <u>0.7046</u> | 0.6975 | 0.6948 | <u>0.6937</u> | 0.6910 | 0.6811 |
| Gemma3-27B-IT | <u>0.6968</u> | 0.6857 | 0.6688 | <u>0.5616</u> | 0.5454 | 0.5396 |

Table 32: Task 2 accuracy for English conversations across approaches and complex instruction formats. The best result per column is **bolded** and the best result per model-approach pair is <u>underlined</u>.

| Model | $f_1$ | $f_2$ | $f_3$ |
|---|---|---|---|
| Claude3.7-Sonnet | 0.9529 | **0.9574** | **0.9565** |
| Claude3.5-Sonnet-V2 | <u>0.9532</u> | 0.9498 | 0.9471 |
| Claude3.5-Sonnet-V1 | 0.9254 | <u>0.9356</u> | 0.9251 |
| Gemma3-27B-IT | <u>0.8952</u> | 0.8734 | 0.8695 |
| Qwen3-14B | <u>0.2269</u> | 0.2208 | 0.2018 |
| Llama3.3-70B | 0.1435 | <u>0.1610</u> | 0.1511 |

Table 33: Task 3 compliance rate across models and complex instruction formats (only English conversations). The best result per complex instruction format is **bolded** and the best result per model is <u>underlined</u>.